\documentclass[letterpaper, 10 pt, conference]{ieeeconf}  

\IEEEoverridecommandlockouts                              

\usepackage{graphics} 
\usepackage{epsfig} 
\usepackage{times} 
\usepackage{amsmath} 
\usepackage{amssymb}  
\usepackage{algorithm}
\usepackage{algpseudocode}

\usepackage{color}
\usepackage{xcolor}

\usepackage{xurl}

\title{\LARGE \bf
Multi-embodiment Legged Robot Control as a Sequence Modeling Problem
}

\author{Chen Yu$^{1,2}$, Weinan Zhang$^{3}$, Hang Lai$^{2,3}$, Zheng Tian$^{4}$, Laurent Kneip$^{1}$, and Jun Wang$^{2,5}$
\thanks{$^{1}$School of Info. Sci. and Tech., ShanghaiTech University, China.}
\thanks{$^{2}$Digital Brain Lab, Shanghai, China}%
\thanks{$^{3}$Dept. of Computer Sci. and Eng., Shanghai Jiao Tong University, China.}
\thanks{$^{4}$School of Creativity and Art, ShanghaiTech University, China.}%
\thanks{$^{5}$Centre for Artificial Intelligence, University College London, UK.}
}

\begin{document}

\maketitle
\thispagestyle{empty}
\pagestyle{empty}

\begin{abstract}

Robots are traditionally bounded by a fixed embodiment during their operational lifetime, which limits their ability to adapt to their surroundings.
Co-optimizing control and morphology of a robot, however, is often inefficient due to the complex interplay between the controller and morphology.
In this paper, we propose a learning-based control method that can inherently take morphology into consideration such that once the control policy is trained in the simulator, it can be easily deployed to robots with different embodiments in the real world. In particular, we present the Embodiment-aware Transformer (EAT), an architecture that casts this control problem as conditional sequence modeling. EAT outputs the optimal actions by leveraging a causally masked Transformer. By conditioning an autoregressive model on the desired robot embodiment, past states, and actions, our EAT model can generate future actions that best fit the current robot embodiment.
Experimental results show that EAT can outperform all other alternatives in embodiment-varying tasks, and succeed in an example of real-world evolution tasks: stepping down a stair through updating the morphology alone. We hope that EAT will inspire a new push toward real-world evolution across many domains, where algorithms like EAT can blaze a trail by bridging the field of evolutionary robotics and big data sequence modeling.


\end{abstract}

\section{Introduction}

In nature, animal species can exhibit physiological and structural adaptations to changes in environments across multiple generations. This increases their likelihood of survival and the preservation of their genes \cite{raff2012shape}. However, in the field of robotics---although more and more robots show their ability to evolve their controller through interaction with the real world to improve their adaptivity to the environment \cite{smith2022legged, yu2022learning, ha2020learning, massi2019combining}---real robots are traditionally bounded by a fixed embodiment during their operational lifetime. 


Some previous works in the field of Evolutionary Robotics optimize morphology together with control of robots \cite{eiben2021real, nygaard2021real, nygaard2021environmental, nygaard2020real, nygaard2018real, rosendo2017trade, vujovic2017evolutionary, joachimczak2016artificial}. These robots and controllers are relatively simple and hence hard to be deployed in real applications. Other works propose hierarchical approaches with two loops: The outer loop evolves morphology while the inner loop optimizes a controller for each new morphology \cite{le2022morpho, goff2021challenges, zhao2020robogrammar, luck2020data, jelisavcic2019lamarckian}. However, for relatively complicated robots that involve dynamic locomotion, hierarchical approaches often only work in simulation, as it would usually take millions of control steps for evolution \cite{gupta2021embodied}.

\begin{figure}[thpb]
  \centering
  \framebox{\parbox{0.47\textwidth}{
  \centering
  \includegraphics[width=0.42\textwidth]{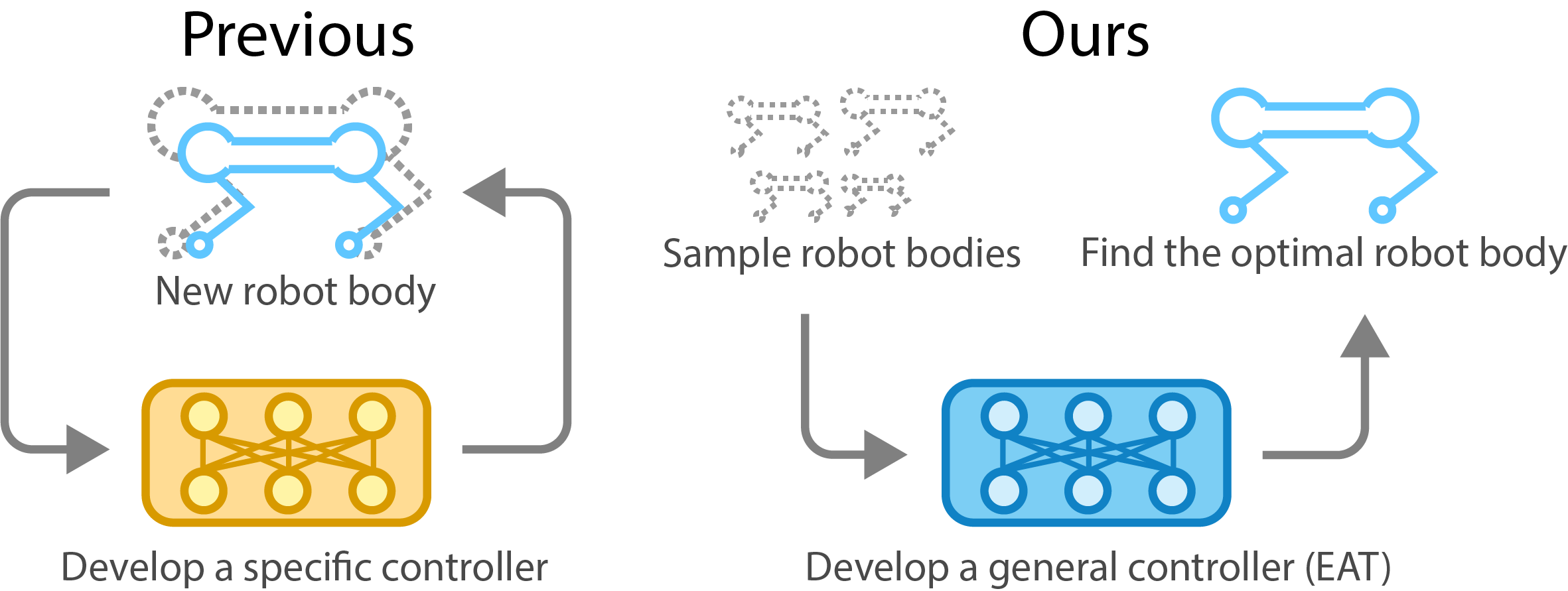}
}}
  \caption{\textbf{Demonstration of two robot evolution schemes.} We compare the hierarchical approaches (left) and our method based on a general purpose controller (right).
}
  \label{fig:comp}
  \vspace{-0.1cm}
\end{figure}

To tackle these challenges, in this work, we wish to have a controller that inherently takes morphology into consideration. In this way, once a general control policy is trained (e.g., in the simulator), it can be deployed on robots with different embodiments (e.g., in the real world), as shown in Fig. \ref{fig:comp}. This is usually challenging because of the complicated interplay between robot morphology and control \cite{picardi2021morphologically}.

While traditional model-based control approaches for robots---especially with relatively complicated dynamics, such as legged robots---are usually based on analytic dynamics models \cite{di2018dynamic, minniti2021adaptive, grandia2019feedback}, it is possible to control such a robot with varying morphology by system identification \cite{park2010system, nagarajan2009state}. However, this requires a considerable amount of tedious hand-engineering and a known robot model.

Learning-based methods, such as reinforcement learning (RL), have proven effective at solving an increasing number of real-world locomotion tasks \cite{lee2020learning, kumar2021rma, miki2022learning}. Chen et al. \cite{chen2018hardware} formulate the policy as a function of the current state and the hardware property encoding. However, it either requires the full kinematics information of the robot, or implicitly learns the hardware representation, in which case it is challenging for zero-shot transfer to unseen robots. Schaff et al. \cite{schaff2019jointly} maintain a distribution over designs and use RL to maximize expected rewards over the design distribution. Since the policy optimization is still embedded in a control-morphology co-optimization pipeline, it would be inefficient to update the morphology in the real world. Regarding the morphology as a graph structure, preliminary works also explore this problem through graph neural networks for morphology generalization \cite{huang2020one, pathak2019learning, wang2018nervenet}, but none of which is validated in the real world.
Concurrently with our work, Feng et al. \cite{feng2022genloco} propose an RL-based general-purpose locomotion controller,
GenLoco, using morphology randomization. They validate their controller on three commercially-available robots in the real world. 




Note there are recent works formulating RL as a sequence modeling problem \cite{srivastava2019training,reed2022generalist}. Decision Transformer \cite{chen2021decision, reid2022can, zheng2022online} uses state, action, and returns-to-go (sum of future rewards) as tokens in a Transformer model. 
Trajectory Transformer \cite{janner2021offline} uses a Transformer model to predict the dynamics of a robot and uses beam search \cite{reddy1977speech} for planning.
These Transformer-based approaches have achieved similar or better performances in benchmark tasks compared with classic RL algorithms thanks to the model capacity and the self-attention mechanism.

In this work, we also take advantage of Transformer models to design a controller for robots with changing morphology.
In particular, we present Embodiment-aware Transformer (EAT), an architecture that casts this control problem as conditional sequence modeling. EAT uses a Transformer architecture to model distributions over trajectories and robot morphology and outputs the optimal actions by leveraging the causally masked Transformer. 
By conditioning the autoregressive model on the desired robot embodiment (we focus on the morphology in this work), past states, and actions, our EAT model can generate future actions that best fit the current robot embodiment, as shown in Fig. \ref{fig:net}.


We validate EAT on a locomotion control task---learning to walk stably from scratch---on the quadruped robot Mini Cheetah \cite{katz2019mini} both in the simulator and in the real world. We allow the robot to grow its body and leg size to simulate ontogenetic morphological changes.
After deploying the trained EAT model on the real robots, we further use Bayesian Optimization (BO) \cite{snoek2012practical} to optimize the morphology of the robots online, while the fixed EAT policy is conditioned on the current body shape.


Experimental results show that EAT can successfully find the controller that fits the current robot morphology. As a consequence, it outperforms all other alternatives in the simulator and succeeds in our real-world evolution task.
To the best of our knowledge, this is the first time that a Transformer is applied to an evolutionary robotics task; and this is the first time that real-world morphology evolution is applied to a dynamic quadruped robot.



\section{Preliminaries}


The Transformer model is introduced by Vaswani et al. \cite{vaswani2017attention} for efficient sequential data modeling, which has been shown to perform strongly on various tasks from Natural Language Processing \cite{kitaev2018constituency, liu2018generating} to Computer Vision \cite{liu2022video, meinhardt2022trackformer}.
It consists of stacked self-attention layers with residual connections.
Each self-attention layer
maps an input sequence of symbol representations $(x_1, \dots, x_n)$ with a context length of $n$ to a sequence
of continuous representations $\mathbf{z}=\left(z_{1}, \ldots, z_{n}\right)$.
Each token is mapped linearly to a key $k_{i}$, query $q_{i}$, and value $v_{i}$. The corresponding output of the self-attention layer is given by weighting the values $v_{j}$ by the normalized dot product between the query $q_{i}$ and other keys $k_{j}$:
\begin{equation}
z_{i}=\sum_{j=1}^{n} \operatorname{softmax}\left(\frac{\left\{\left\langle q_{i}, k_{j^{\prime}}\right\rangle\right\}_{j^{\prime}=1}^{n}}{\sqrt{d_k}}\right)_{j} \cdot v_{j},
\end{equation}
where $d_k$ is the dimension of queries and keys.

This can be used for forming state-return associations via similarity of the query and key vectors in the context of offline RL \cite{chen2021decision}.
Offline RL algorithms learn effective policies from previously collected, fixed datasets without further environment interaction \cite{ernst2005tree, fujimoto2019off, kumar2019stabilizing}.



\section{Embodiment-aware Transformer}
\label{sec:eat}
\begin{figure}[thpb]
  \centering
  \framebox{\parbox{0.47\textwidth}{
  \includegraphics[width=0.47\textwidth]{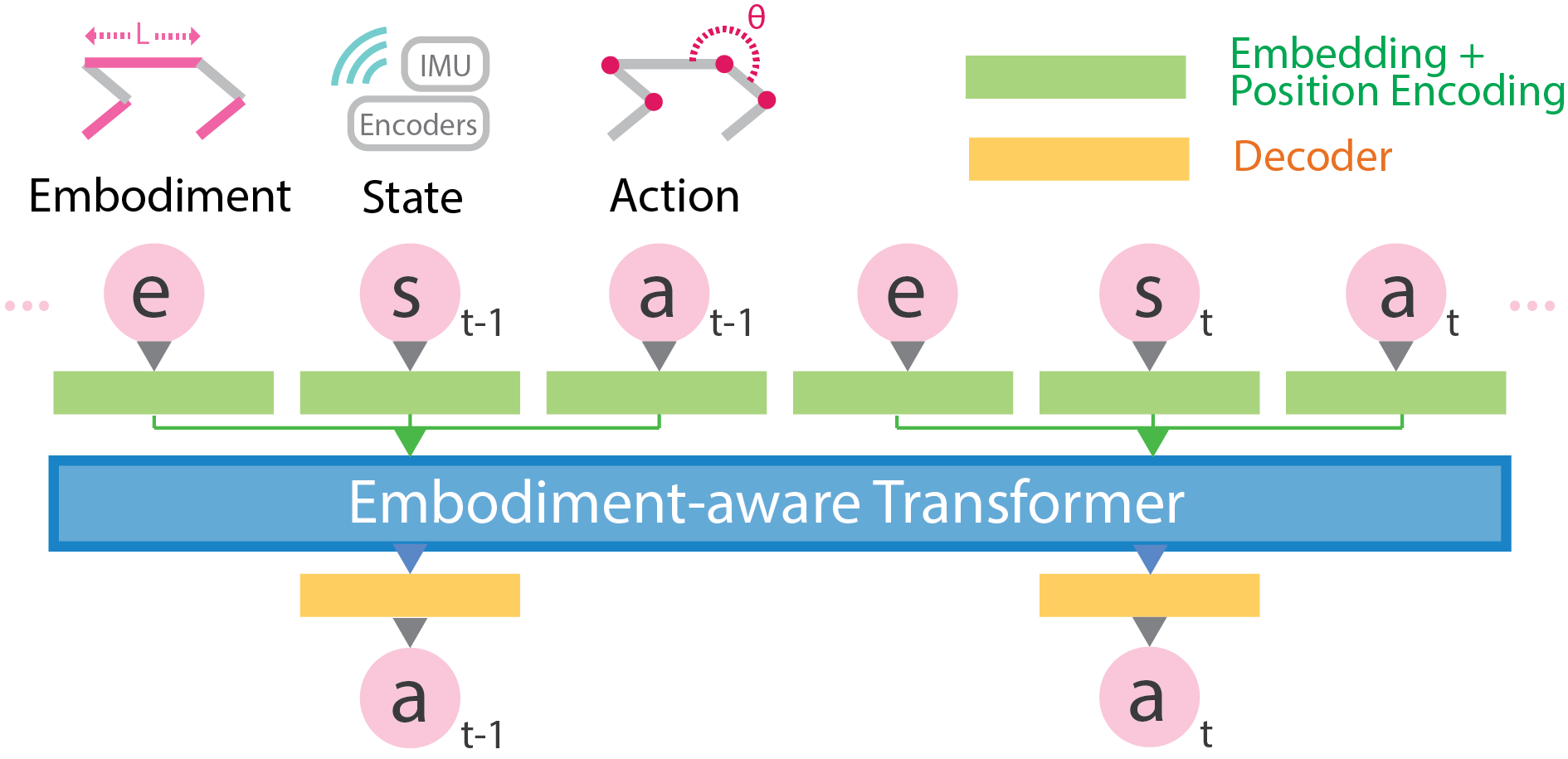}
}}
  \caption{\textbf{Embodiment-aware Transformer architecture.} We learn a linear layer for embodiment, states, and actions for token embeddings, while a positional episodic timestep encoding is added. Tokens are fed into a GPT model that predicts actions autoregressively with a causal self-attention mask.
}
  \label{fig:net}
\end{figure}

In this section, we present Embodiment-aware Transformer,
as summarized in Fig. \ref{fig:net} and Algorithm \ref{alg:eat}.

\subsection{Embodiment-aware Markov Decision Process}
We model the control of the robot as a variant of Markov decision process (MDP), referred to as embodiment-aware MDP, described by the tuple $(\mathcal{E}, \mathcal{S}, \mathcal{A}, P_E, \mathcal{R})$. The MDP tuple consists of embodiment $e \in \mathcal{E}$, states $s \in \mathcal{S}$, actions $a \in \mathcal{A}$, embodiment-dependent state transition dynamics $P_E\left(\cdot \mid s, a; e\right)$, and a reward function $r=\mathcal{R}(s, a)$.
We use $e$ to denote the robot embodiment and $s_{t}, a_{t}$, and $r_{t}$ for state, action, and reward at timestep $t$, respectively. Within each episode, we sample an embodiment $e$ from a distribution $\rho(e)$. 
Given a policy $\pi(\cdot \mid s, e)$, a trajectory can be generated by interacting with the environment, which is made up of a sampled embodiment and a sequence of states, actions, and rewards: $\tau=\left(e, s_{1}, a_{1}, r_{1}, s_{2}, a_{2}, r_{2}, \ldots, s_{T}, a_{T}, r_{T}\right)$, where $T$ is the episode length.
Similar to the setting of standard MDP, our goal is to learn a policy that maximizes the expected return across different embodiment-aware MDPs:
\begin{equation}
  \pi^* := \mathop{\arg \max}_\pi \mathbb{E}_{\rho, \pi, P_E}\Big[\sum_{t=1}^{T} r_{t}\Big].
\end{equation}
We set the reward discount factor as $1$ here since we assume each step is equally important.
\vspace{5pt}
\subsection{Embodiment-aware Transformer}

\noindent\textbf{Trajectory representation.} Different from Decision Transformer \cite{chen2021decision}, we expect the trajectory representation to enable the Transformer to learn meaningful patterns between robot embodiment and actions, and the Transformer to conditionally
generate actions based on embodiment at test time. Therefore, instead of feeding the returns-to-go as in \cite{chen2021decision}, we feed the vector representation $e$ of the robot embodiment (e.g., length of legs and torso of the robot) at each timestep. This leads to the following trajectory representation which enables autoregressive training and generation:
\begin{equation}
    \tau = (e, s_1, a_1, e, s_2, a_2,\dots, e, s_{T}, a_{T}).\label{eq:trajectory}
\end{equation}

Note that although theoretically, the robot embodiment $e$ in Eq.~(\ref{eq:trajectory}) can be time-varying, it is set as fixed during an episode in our experiment since varying the body shape of the real robot within an episode is impractical so far.

\vspace{5pt}
\noindent\textbf{Training.}
For robots with $M$ types of sampled embodiment, we are given a dataset $D_{\text{expert}}^i$ of expert demonstration of offline trajectories $\tau$ for each of the $i$-th embodiment. We integrate these datasets and feed the sampled $H$ timesteps of each trajectory into EAT.
For token embeddings, we learn linear embedding layers for robot embodiment $e$, state $s$, and action $a$, with layer normalization \cite{ba2016layer}. Similar to \cite{chen2021decision}, an embedding for each timestep is added to each token. Here, such a tuple $(e, s, a)$ plays a similar role as a ``word'' in a language model.
The tokens are then processed by a Generative Pre-trained Transformer (GPT) \cite{radford2018improving} model, which autoregressively predicts future tokens tuples $(e, s, a)$. The predicted action is used for calculating the mean-squared error for backpropagation.


\vspace{5pt}
\noindent\textbf{Evaluation.}
For evaluation, we can specify the current robot embodiment $e$ and the robot's initial state, as the conditioning information to initiate generation. We feed the last $H$ timesteps of the current trajectory $\tau$ into EAT to obtain the predicted action for the last observed state.

\begin{algorithm}[t]
\caption{Embodiment-aware Transformer (EAT)}
\begin{algorithmic}[1]

\Require $D_{\text{expert}}^0, D_{\text{expert}}^1,\dots, D_{\text{expert}}^{M-1}$
\Ensure Trained EAT Model
\State $D_{\text{expert}} = D_{\text{expert}}^0 \cup D_{\text{expert}}^1 \cup \dots \cup D_{\text{expert}}^{M-1}$
\For{$i = 0, \dots, I - 1 $ iterations}
\State Sample $H$-long ($e, s, a, t$) in $D_{\text{expert}}$
\State Stack embeddings of ($e, s, a$) for each timestep
\State Feed the Stacks to the GPT model
\State Update the GPT model
\EndFor
\end{algorithmic}
\label{alg:eat}
\end{algorithm}

\subsection{Evolution of Morphology with EAT}

Once we have a unified control policy $\pi_{\text{EAT}}$ for robots with varying morphology, we can further apply EAT to a real-world evolution task to find the best morphology:
\begin{equation}
e^\star=\underset{e \in \mathcal{E}}{\operatorname{argmax}} f(e | \pi_{\text{EAT}}),
\end{equation}
where the objective $f: \mathcal{E} \rightarrow \mathbb{R}$ is a black-box function that can be different from the reward function $r$. We can use an off-the-shelf black-box optimizer, such as BO, to solve this problem, as shown in Algorithm \ref{alg:bo}.

\begin{algorithm}[t]
\caption{Real-world Evolution}
\begin{algorithmic}[1]
\Require Trained EAT Model $\pi_{\text{EAT}}$, $m_0$
\Ensure $e^\star$
\For{$g = 0, \dots, G - 1 $ generations}
\For{$t = 0, \dots, T - 1 $ timesteps}
\State Action = $\pi_{\text{EAT}}(e_g, s_t, a_t, t)$
\EndFor
\State Optimizer give the next candidate $m_g$
\EndFor
\State \textbf{return} The best morphology $e^\star$.
\end{algorithmic}
\label{alg:bo}
\end{algorithm}

In this way, we solve the inefficiency challenge of the morphology-controller co-optimization problem by taking advantage of our embodiment-aware control mechanism.

\section{Evaluations in Simulation}

\begin{figure}[thpb]
  \centering
  \framebox{\parbox{0.47\textwidth}{
  \includegraphics[width=0.47\textwidth]{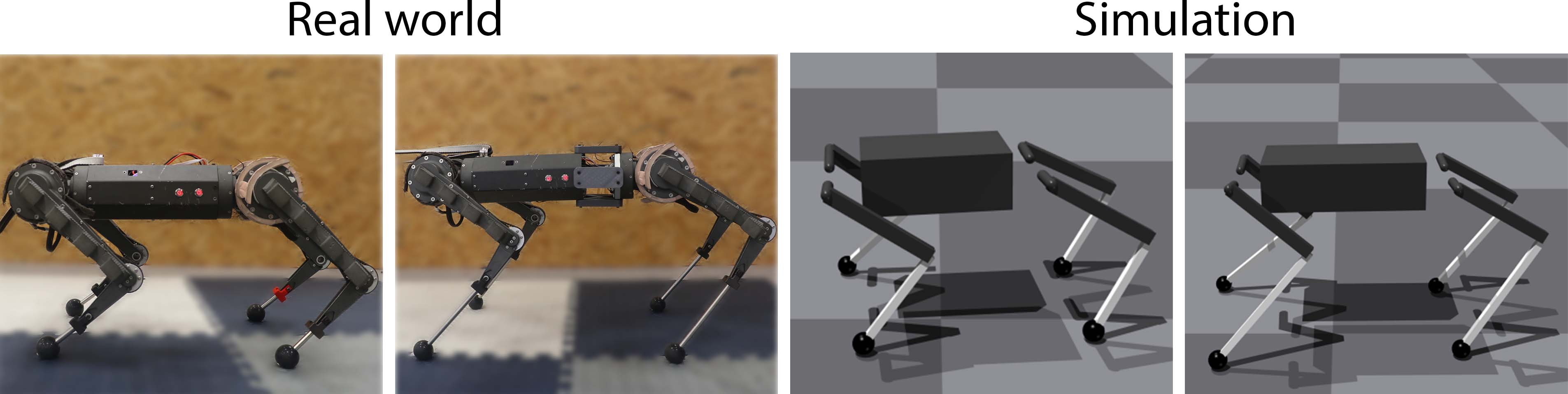}
}}
  \caption{\textbf{The robot with changeable morphology.} We conduct the real-world experiment on a quadruped robot with variable lengths of lower limbs and torso. We show the robot with embodiment representation $e$ of (0.27 m, 0.2 m, 0.2 m) and (0.35 m, 0.3 m, 0.3 m) on the top and the corresponding simulated counterparts on the bottom.
}
  \label{fig:photos}
\end{figure}

\begin{figure*}[thpb]
  \centering
  \framebox{\parbox{0.98\textwidth}{
  \includegraphics[width=0.98\textwidth]{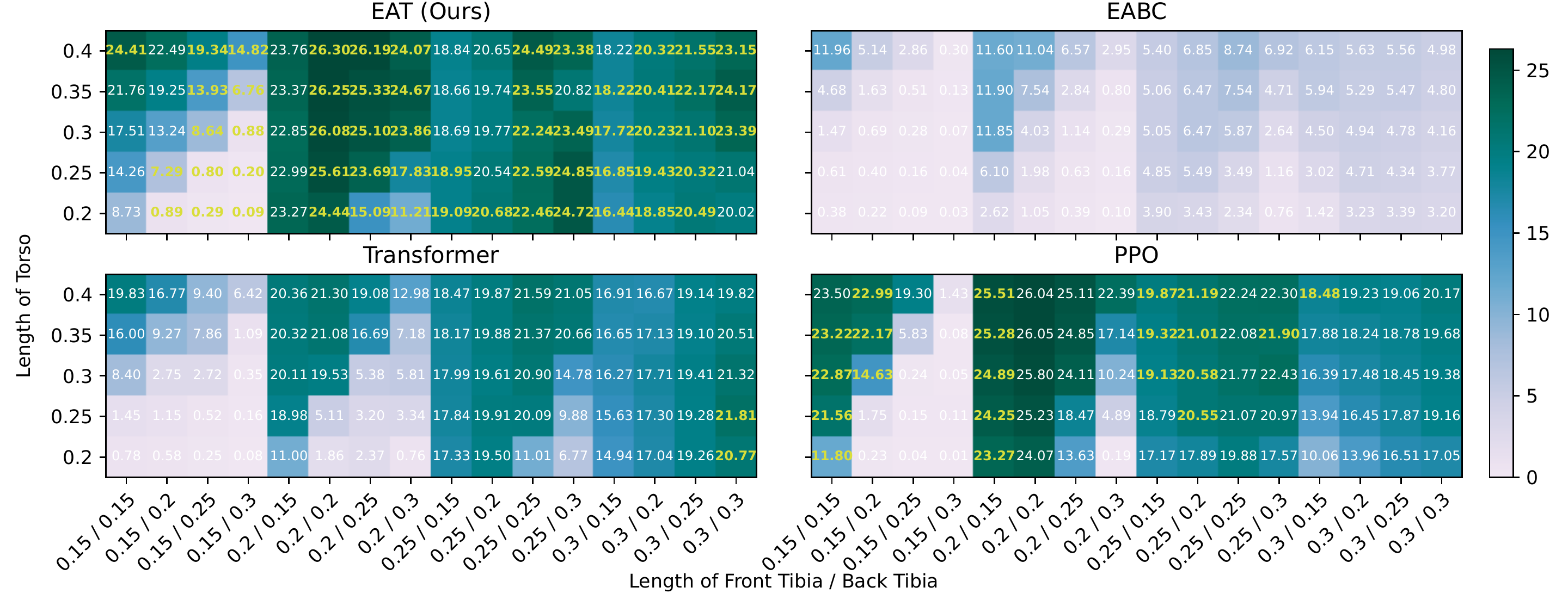}
}}
  
  \caption{\textbf{Performance of trained policies on robots with different embodiments for a locomotion task of walking on a flat plane.} Each block corresponds to a specific robot embodiment---encoded by the length of torso, front tibia, and hind tibia---whose color represents the accumulated reward within an episode using a single trained policy. The best score for each robot embodiment is highlighted in yellow. In most cases, EAT can have the best performance compared with other alternatives. 
}
  \label{fig:sim}
  \vspace{-0.3cm}
\end{figure*}

In this section, we focus on the locomotion training tasks of the quadruped robot Mini Cheetah \cite{katz2019mini} with variable lengths of the tibia and torso in simulation (Fig. \ref{fig:photos}).
We want to train the robot to walk stably from scratch using EAT as described in Section \ref{sec:eat}.
The setup of this task is similar to \cite{rudin2022learning}.
Specifically, the state $s$ here consists of: base linear and angular velocities, the gravity vector, joint positions and velocities, and the previous actions performed by the policy. The action $a$ is the desired joint positions of the motors, sent to a PD controller. The total reward is a weighted sum of nine terms, as in \cite{makoviychuk2021isaac}, including velocity tracking error, velocity penalty, action penalty, and a bonus of feet-in-the-air duration. We defer the reward details to \cite{rudin2022learning}.

The embodiment vector $e$ here consists of three dimensions: front tibia length, hind tibia length, and torso length. We sample some robots with specific embodiment $e_i$: 0.2, 0.25, and 0.3 m for front tibia length; 0.2, 0.25, and 0.3 m for hind tibia length; 0.2, 0.3, and 0.4 m for torso length.
We use Proximal Policy Optimization (PPO) to train these 27 ($3\times3\times3$) robots separately with domain randomization \cite{tan2018sim} in Isaac Gym \cite{makoviychuk2021isaac}. For all embodiment, we keep the reward function and initial state consistent. We collect 1000 trajectories from each environment for the training of EAT, while each trajectory has a length of 1000 timesteps.


We compare EAT with
Embodiment-aware Behaviour Cloning (EABC), Vanilla Transformer, and a single PPO policy. For EABC, we concatenate the embodiment vector with the robot observation; for Vanilla Transformer, we use the same model architecture as EAT but without the embodiment vector $e$ in the trajectory representation. We use the same training dataset as for EAT for these two methods. We also evaluate a baseline method:
run a single PPO policy---that trained on the original shape of Mini Cheetah---directly on robots with variable embodiments. Experimental results are shown in Fig. \ref{fig:sim}, where each block corresponds to a specific robot embodiment and the color represents the accumulated reward within an episode using the same trained policy. For each robot, we evaluate the performance of 1000 agents in parallel for 3 trails and report the average results.

A single EAT policy can successfully control robots with different shapes to walk: Except in some cases where robots have extremely unbalanced shapes, EAT can achieve a return higher than 10 in 70 / 80 cases.
For 57 / 80 of the robot embodiments, EAT can outperform all other methods (scores highlighted in yellow). Specifically, for body shapes that are not in the training dataset (body length of 0.25 m and 0.35 m; front tibia and hind tibia of 0.15 m), EAT can still have a good walking score. For example, for the zero-shot evaluation of embodiment with a torso length of 0.4 m (the first line of the return matrix), EAT achieves the best score among these methods in 11 / 16 cases.
This shows the generalization ability of EAT for unseen embodiments.
EAT outperforms all other methods on average (Tab. \ref{tab:scores}). 

To evaluate the robustness of EAT, we add noises to the evaluation environment, including random friction coefficient, base mass, and pushing. The amplitudes of these noises are twice that used in the training stage. Results are shown in Tab. \ref{tab:scores}, where EAT demonstrates its superiority even in a noisy environment. We will show in our real-world experiment that Transformer-based controllers are more robust than PPO, especially after the sim-to-real transfer.

\begin{table}
\caption{Scores of locomotion using different methods.}
\begin{center}
\begin{tabular}{|c|c|c|}
\hline
\textbf{Method}& \textbf{Average Score}& \textbf{Average Score in Noisy Env.}  \\
\hline
EAT (Ours)\rule{0pt}{2ex} & $\mathbf{18.87 \pm 0.52}$ & $\mathbf{18.61 \pm 0.13}$\\
EABC & ${3.8 \pm 0.9}$ & ${2.18 \pm 1.32}$\\
Transformer & ${13.32 \pm 1.37}$ & ${13.49 \pm 1.38}$\\
PPO & ${16.53 \pm 0.57}$ & ${14.26 \pm 1.01}$\\
\hline
EAT with $H=1$ \rule{0pt}{2ex}  & ${15.06 \pm 0.53}$ & ${14.15 \pm 0.32}$ \\
\hline
EAT with LD \rule{0pt}{2ex} & ${13.54 \pm 0.16}$ & ${13.64 \pm 1.19}$ \\
EAT with LD-5k & ${14.19 \pm 0.55}$ & ${14.07 \pm 0.67}$ \\
EAT with LD-10k & ${15.32 \pm 1.18}$ & ${15.35 \pm 0.34}$ \\
\hline
\end{tabular}
\end{center}
\label{tab:scores}
\vspace{-0.3cm}
\end{table}


We hypothesize that the superiority of EAT comes from the context information of previous tokens and the capacity of the model for fitting our diverse training dataset that contains both varying trajectories and robot morphology.
To investigate the importance of access to previous states and actions, we ablate on the context length $H$. 

The performance of EAT degrades significantly when the horizon length is 1 (Tab. \ref{tab:scores}), indicating that past information is essential for this morphology-varying locomotion task. 

\begin{figure*}[thpb]
  \centering
  \framebox{\parbox{0.98\textwidth}{
  \includegraphics[width=0.98\textwidth]{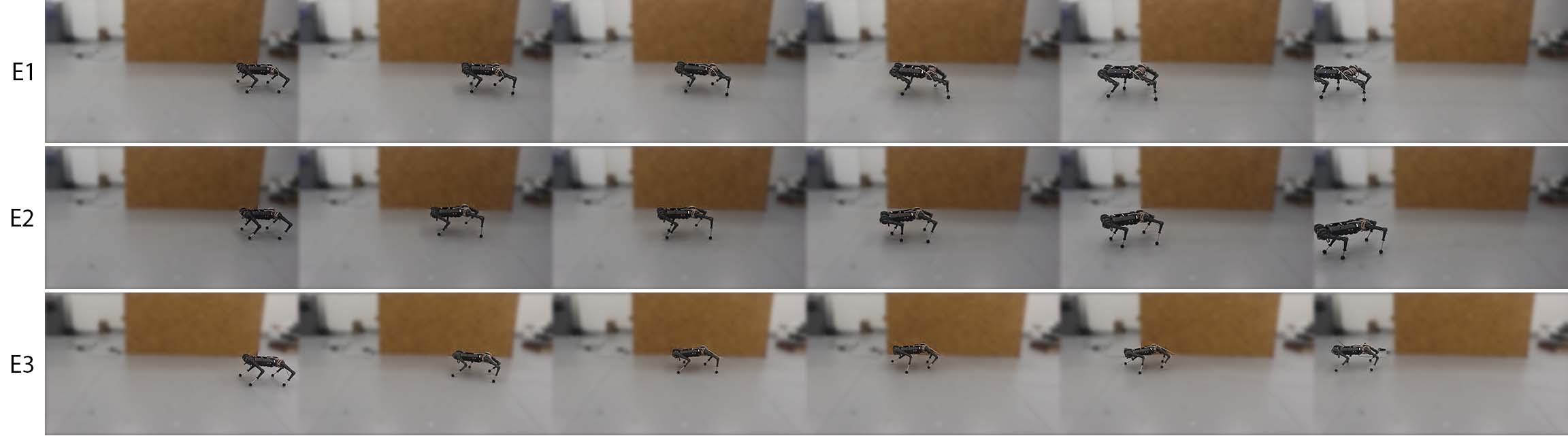}
}}
  
  \caption{\textbf{Robots with different morphologies controlled by the same EAT policy.} We run a trained EAT policy on robots with embodiment (0.3 m, 0.2 m, 0.2 m), (0.3 m, 0.2 m, 0.25 m), and (0.3 m, 0.25 m, 0.2 m) in each row of the figure respectively. EAT can successfully control the robot to move forward in these three cases where the positions of CoM are different.
}
  \label{fig:plane}
\end{figure*}
\begin{figure*}[thpb]
  \centering
  \framebox{\parbox{0.98\textwidth}{
  \includegraphics[width=0.98\textwidth]{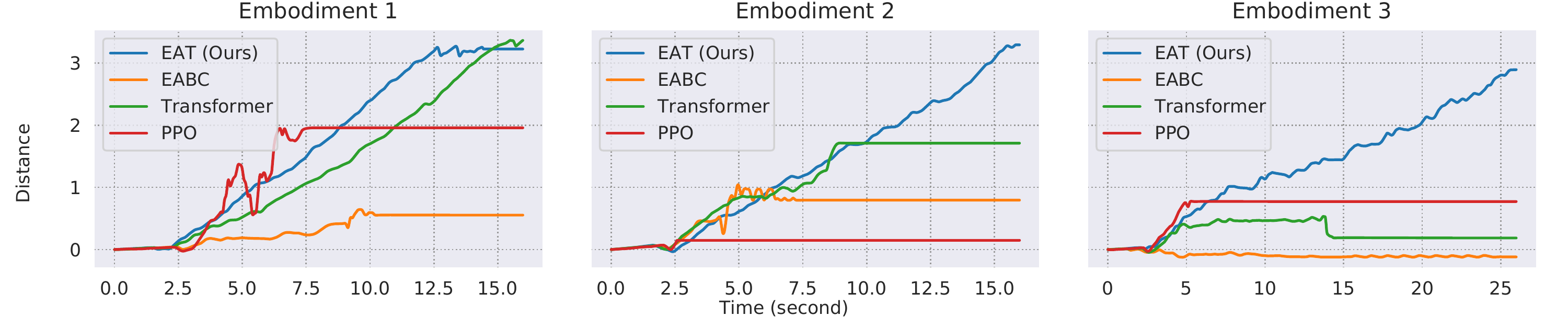}
}}
  
  \caption{\textbf{Walking trajectories of robots with three different embodiments.} Embodiment 1 represents a body shape of the original Mini Cheetah, where the front lower limbs and hind lower limbs have the same medium length (0.2 m). In this case, both EAT and Vanilla Transformer can enable the robot to move forward, while PPO leads the robot to unstable states because of the reality gap. Embodiments 2 and 3 represent robots with asymmetrical front and hind lower limbs, leading to forward and backward CoM respectively. In these two cases, only EAT can successfully control the robot.
}
  \label{fig:track}
  \vspace{-0.3cm}
\end{figure*}

We further investigate the impact of the training dataset on the performance of EAT.
We consider another training dataset that contains only 8 robot shapes instead of 27: torso length is 0.2 m or 0.4 m; front and back tibia length is 0.2 m or 0.3 m. We refer to this dataset as LD (Less Diverse). We also test on training datasets with variants of LD: LD with 5k trajectories per robot morphology and a 10k version, denoted as LD-5k and LD-10k.

Results in Tab. \ref{tab:scores} show that training datasets with fewer types of robots (dataset LD) yield degraded performance of EAT, even if the dataset is larger (dataset LD-5k and LD-10k). This suggests that EAT can leverage the embodiment information in the dataset and implicitly build state-return associations via the similarity of the query and key vectors. This makes it superior in embodiment-conditioned control. 

\section{Real-world Experiment}
\label{sec:exp}
We directly deploy the EAT model that is trained in the simulator to the real robots without any fine-tuning. The representation of the state, action, and embodiment remains the same as in simulation.
We change the originally fixed tibias of the robot Mini Cheetah to configurable modules that we can easily adjust the length by replacing the steel tube, and design 3D-printed parts to lengthen the torso, as shown in Fig. \ref{fig:photos}, with an accuracy of 1 cm. 

\subsection{EAT for Walking on Plane}
We start from the same scenario as in the simulators: walking on real-world flat terrain. We do experiments on three different embodiments: $(0.3 m, 0.2 m, 0.2 m)$, $(0.3 m, 0.2 m, 0.25 m)$, and $(0.3 m, 0.25 m, 0.2 m)$, referred to Embodiment 1, 2, and 3. These embodiments for testing represent three cases: 1) robots with the shape of the original Mini Cheetah; 2) robots with an asymmetric shape and a forward Center of Mass (CoM); and 3) robots with a backward CoM. However, our model is not limited by only these three embodiments. Rather, we use them as representative examples for demonstration purposes, and we expect that EAT can generalize to other possible embodiments.

Fig. \ref{fig:plane} snapshots of successful walking with these three embodiments using EAT and Fig. \ref{fig:track} shows the walking distance in the first 15 or 25 seconds.

\begin{figure*}[thpb]
  \centering
  \framebox{\parbox{0.98\textwidth}{
  \includegraphics[width=0.98\textwidth]{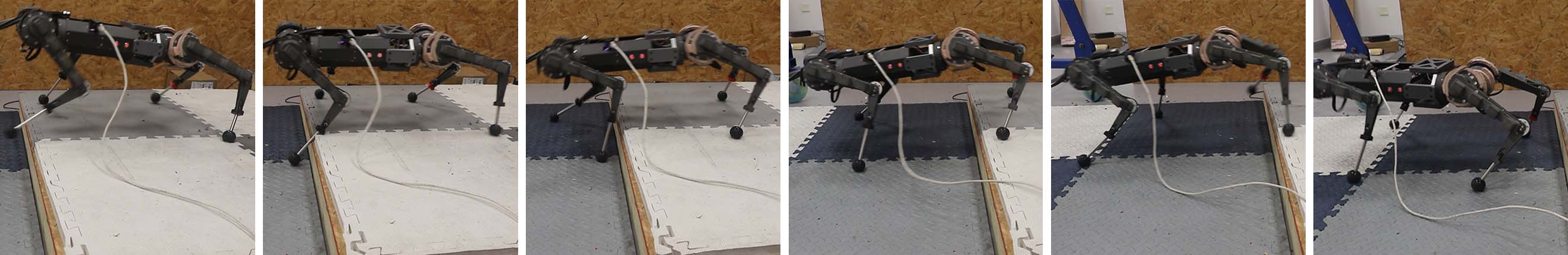}
}}
  
  \caption{\textbf{Snapshots of the robot waking down a stair after morphology evolution.} Our robot with the EAT controller successfully find a morphology that suits this example task of stepping down: It finds a morphology of shorter front tibias (0.19 m), longer back tibias (0.22 m), and a longer torso (0.35 m). EAT successfully control the robot with this morphology, and this morphology helps the robot to keep balance during stepping down.
}
  \label{fig:stair}
  \vspace{-0.3cm}
\end{figure*}

Experimental results of Embodiment 1 reveal the sim-to-real transfer ability of different control methods. In this test, EAT and Vanilla Transformer successfully controls the robot to walk stably. Robots with PPO can walk for the beginning few steps but quickly goes into unstable states as we can see in the accompanying video and Fig. \ref{fig:track}.
This is likely because of the reality gap that comes from the mismatch of motor models, floor material, and control latency. These reality gaps continuously feed out-of-distribution (OOD) observations---that do not conform to the underlying distribution of the training data---to the policy.
Thanks to the generalization ability of Transformers when decoding and the stable nature of offline RL, both EAT and Vanilla Transformer can overcome this reality gap and perform as in the simulator. 

For Embodiment 2, PPO still leads the robot to unstable states very quickly. Robots with Vanilla Transformer fall forward after the first few steps, because of the unstable CoM of the robot. EAT successfully finds a control strategy to balance the robot and walk forward. 

The backward CoM of Embodiment 3 makes the task harder since this would prevent the friction force to pull the robot forward. In this case, PPO still quickly leads the robots to unstable states, and only EAT can control the robot to stably walk forward.


\subsection{EAT for Real-world Evolution}

In this section, we showcase the application of EAT in Evolutionary Robotics: a robot can solve an unseen task by updating the morphology alone, using a fixed control policy.
Here, we consider a locomotion task of walking down a stair of 10 cm, using the EAT policy that is trained in the simulator based only on training data of walking on flat terrains.


We evolve the robot morphology by optimizing the embodiment vector $e$: length of the torso, front lower limb, and back lower limb. For each generation, we use BO to sample a morphology $e_g$ from ranges of (0.27 m, 0.35 m), (0.15 m, 0.25 m), and (0.15 m, 0.25 m) respectively. We feed $e_g$ to the EAT policy and test the walking performance---the furthest distance $f(e_g)$ that the robot can walk across the step---directly in the real world. We use BO to optimize $f(e_g)$ as regard to $e_g$ for a maximum of 20 iterations. The furthest distance $f(e_g)$ is measured from the step to where the robot falls over. We assume that the robot can continuously walk on a flat plane once it walks for a distance of 1.5 m.

\begin{figure}[thpb]
  \centering
  \framebox{\parbox{0.47\textwidth}{
  \includegraphics[width=0.47\textwidth]{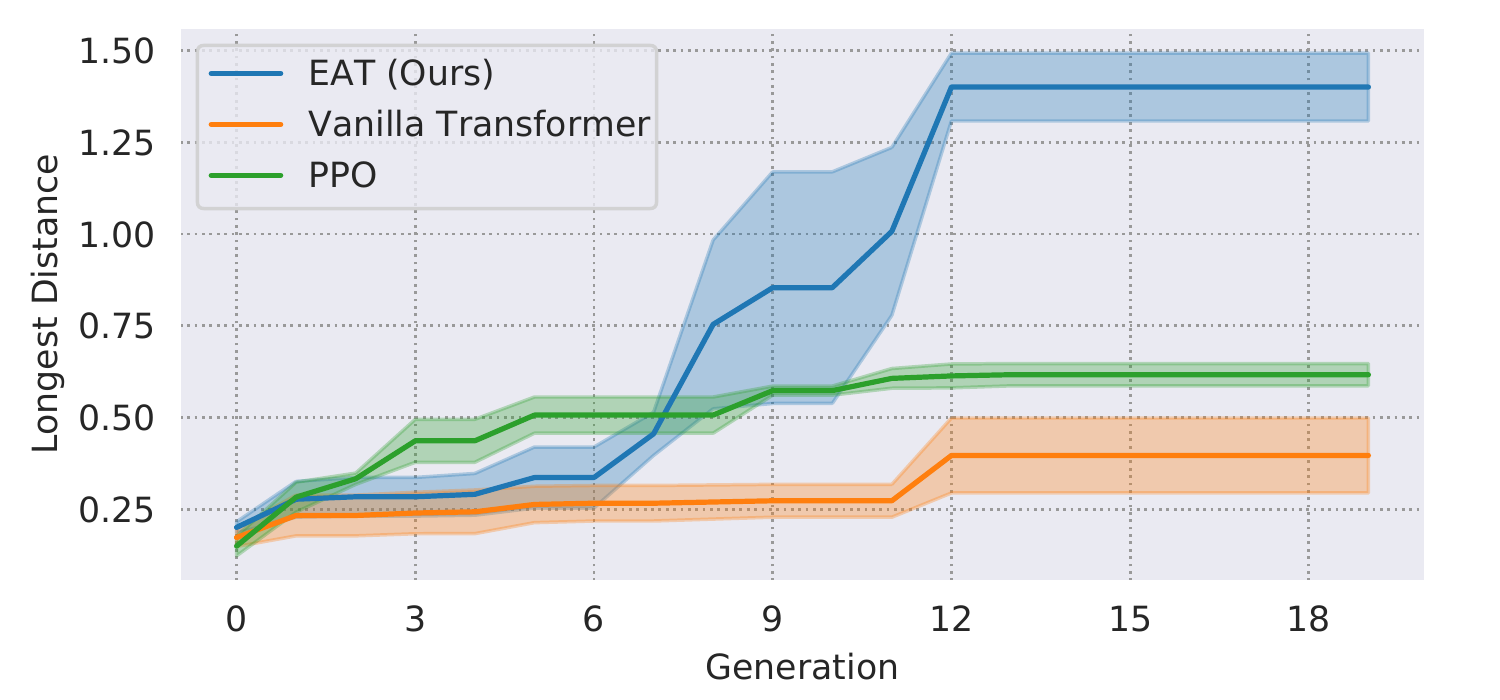}
}}
  \caption{\textbf{Training curves of real-world evolution.} We compare the training processes using the same online morphology optimizer but different backend controllers. Since both PPO and Vanilla Transformer are sensitive to changes in robot morphology, only EAT can succeed in this task: The robot finds a morphology that can adapt to a terrain that is unseen beforehand.
}
  \label{fig:evolve}
  \vspace{-0.3cm}
\end{figure}

Fig. \ref{fig:evolve} compares the training curves when using the three strongest methods---EAT, Vanilla Transformer, and PPO---as the backend controllers while the morphology of the robot is updating. 

EAT together with BO can successfully find the morphology that is capable of solving this task after around 10 iterations: The robot torso is extended to 0.35 m for increasing stability; the length of front tibias is lengthened at 0.22 m and that of hind tibias is shortened at 0.19 m, to provide a backward-oriented CoM that may help to keep balance while walking downstairs (Fig. \ref{fig:stair}).

Consistent with the previous experimental results, as PPO and Vanilla Transformer are sensitive to the morphology changes of our robots, they struggle in this real-world evolution task. Since it is hard for these methods to control a robot with variable morphology using a single policy, they keep falling when a new body shape is given, and hence keep giving the morphology optimizer noisy fitting values.

\section{Discussion}
Eiben \cite{eiben2021real} summarizes two limitations of the current state of the art in Evolutionary Robotics: 1) most studies rely on simulation for reproduction and evolution; 2) robot designs are usually very simple and driven by elementary open-ended control mechanisms. In this work, we attempt to solve these challenges by bridging the fields of big sequential data modeling \cite{reed2022generalist} and evolvable hardware \cite{greenwood2006introduction} for the first time. 


Our proposed evolution pipeline is validated on a task of transversing stairs. Although some previous works have shown the ability of legged robots to climb stairs \cite{shi2022reinforcement, kumar2021rma, siekmann2021blind}, our method goes in a different direction toward embodied intelligence: Unlike most of the previous work, our robot does not see any rough terrain in the simulation training phase, but use only online morphology optimization to overcome the unseen tasks. This unlocks the full potential for evolutionary robots to adapt to an unknown environment that is hard to predict or model beforehand.

Last, although the embodiment representation in our experiment involves only a three-dimension vector, a more sophisticated representation of robot shape---for example, pixel-based representation, which has been successfully embedded in a Transformer network in previous work \cite{chen2021decision, liu2022video, meinhardt2022trackformer}---can also be used in our framework. This inspires more real-world evolution applications involving more diverse morphologies in the future.



\section{CONCLUSIONS}

In this work, we propose Embodiment-aware Transformer (EAT), a learning-based control method that takes actions conditioned on both state and embodiment. We pose this challenge as a conditional sequence modeling problem, and use Transformer to fit a data sequence of (embodiment, state, action) tuples. We train this model using trajectory data generated by multiple PPO policies from various morphologies. Evaluation results on a quadruped robot show that EAT outperforms all other alternatives in cross-embodiment tasks. By leveraging this feature, we apply EAT to a real-world evolution task, i.e., the robot updates its morphology alone online in the real world, for adapting to a locomotion task where the terrain is unseen beforehand.
The success of EAT in solving this task shows the potential of embodiment-aware controller and Transformer in the application of Evolutionary Robotics. We hope that EAT can inspire a new push toward real-world robot evolution based on the recent advance in deep learning-based control.

\newpage









\addtolength{\textheight}{-16.5cm}   

\bibliographystyle{ieeetr}
\bibliography{ref}

\end{document}